\documentclass[10pt,a4paper,conference]{IEEEtran}
\usepackage{cite}
\ifCLASSINFOpdf
  \usepackage[pdftex]{graphicx}
  \graphicspath{{./images/}}
  \DeclareGraphicsExtensions{.pdf,.jpeg,.png}
\else
\fi
\usepackage{amsmath}
\usepackage{amsmath}
\usepackage{amsfonts}
\usepackage{multirow}

\usepackage{authblk}

\begin{document}
\title{Multi-Modal Three-Stream Network for Action Recognition}

% \author{\IEEEauthorblockN{Muhammad Usman Khalid, Fernando Moya \\and Gernot A. Fink}
% \IEEEauthorblockA{Department of Computer Science\\
% TU Dortmund, Germany\\
% usman.khalid@tu-dortmund.de}
% \and
% \IEEEauthorblockN{Jie Yu}
% \IEEEauthorblockA{Corporate Research Department\\
% Bosch, Hildesheim, Germany\\
% jie.yu@de.bosch.com}
% }

\author[1,2]{Muhammad Usman Khalid\thanks{A.A@university.edu}}
\author[1]{Jie Yu\thanks{B.B@university.edu}}
%\author[2]{Fernando Moya\thanks{C.C@university.edu}}
%\author[2]{Gernot A. Fink\thanks{D.D@university.edu}}
\affil[1]{Computer Vision Research Lab, Robert Bosch GmbH, Hildesheim, Germany}
\affil[2]{Department of Computer Science, TU Dortmund, Germany}

\renewcommand\Authands{ and }


%Telephone: (800) 555--1212, Fax: (888) 555--1212}


\maketitle

\begin{abstract}

Human action recognition in video is an active yet challenging research topic due to high variation and complexity of data.
In this paper, a novel video based action recognition framework utilizing complementary cues is proposed to handle this complex problem. Inspired by the successful two stream networks for action classification, additional pose features are studied and fused to enhance understanding of human action in a more abstract and semantic way. 
Towards practices, not only ground truth poses but also noisy estimated poses are incorporated in the framework with our proposed pre-processing module. 
The whole framework and each cue are evaluated on varied benchmarking datasets as JHMDB, sub-JHMDB and Penn Action. 
Our results outperform state-of-the-art performance on these datasets and show the strength of complementary cues.

\end{abstract}


% \fi
\IEEEpeerreviewmaketitle

\section{Introduction}
Human action recognition in video has attracted a lot of attention in varied application domains like autonomous driving, human-machine interaction, video surveillance and health support.
It aims to understand human behavior and interaction by exploiting visual features and temporal dynamics from video. 
One of the major challenges of action recognition is the large variability in human actions, i.e. humans perform a single action differently or single human carries out each action in many ways. In addition, there are variations due to camera position, camera motion, occlusion and resolution.

Recently, impressive progresses in this area have been achieved \cite{simonyan2014_TSCNARV, wang2013_AAPBAR, feichtenhofer2016_CTSNFVAR, wang2015_ARTPDCD}. Effective feature extraction from large amount of video data has proved to be a very crucial factor. For example, the very successful two stream networks proposed in \cite{simonyan2014_TSCNARV, wang2016temporal} are trained individually on RGB frames and optical flows to extract complementary features, i.e. visual appearance and motion dynamics, which are fused in a late fusion manner.
Nevertheless, the performance of these networks is still significantly affected by quantity and quality of data. Current datasets for action recognition in the community are still relatively limited compared to image classification tasks in the sense of diversity and sample quantity, since datasets are relatively small compared to image classification tasks.
Collecting and annotating video datasets demand high amount of resources and time. 
To this end, human poses as high-level and compact description become an important features, as they show good performance on even relatively small datasets. 
The approach proposed in \cite{baradel2017_PCSTAHAR} orders and encodes human joint poses into $3D$ tensors to train a CNN network, fusing its output with a spatial attention mechanism on RGB videos, where all body joints are computed and used. 

    This paper presents a novel approach for exploiting complementary cues: RGB, optical flows, and human poses as data inputs for training. In particular, an end-to-end CNN framework is proposed to train directly on body joint tensor, which can be derived from GT poses or even noisy pose detections by recent pose estimators, e.g. \cite{cao2017realtime}. For the latter case, practical post processing mechanisms are employed to handle imperfect or missing joint detections in realistic videos. Finally, complementary cues for action recognition, i.e. appearance, optical flow and posture features are analyzed and fused to handle varied action classes. Experiments are performed on the challenging action recognition datasets, namely JHMDB, sub JHMDB and Penn Action, our results outperform state-of-the-art approaches.
   
    The remainder of the paper is organized as follows: in Section \ref{sec:relatedWork}, recent trends of action recognition approaches are briefly reviewed. Our proposed Pose ConvNET is introduced in Section \ref{sec:pose_stream}. The following section describes our fusion schemes to incorporate multi-modal inputs. Experiments and evaluation results are given in Section \ref{sec:experiments}. In the last section, the proposed approaches and results are concluded.

  \begin{figure*}[!ht]
    \centering
    \includegraphics[width=4.3in]{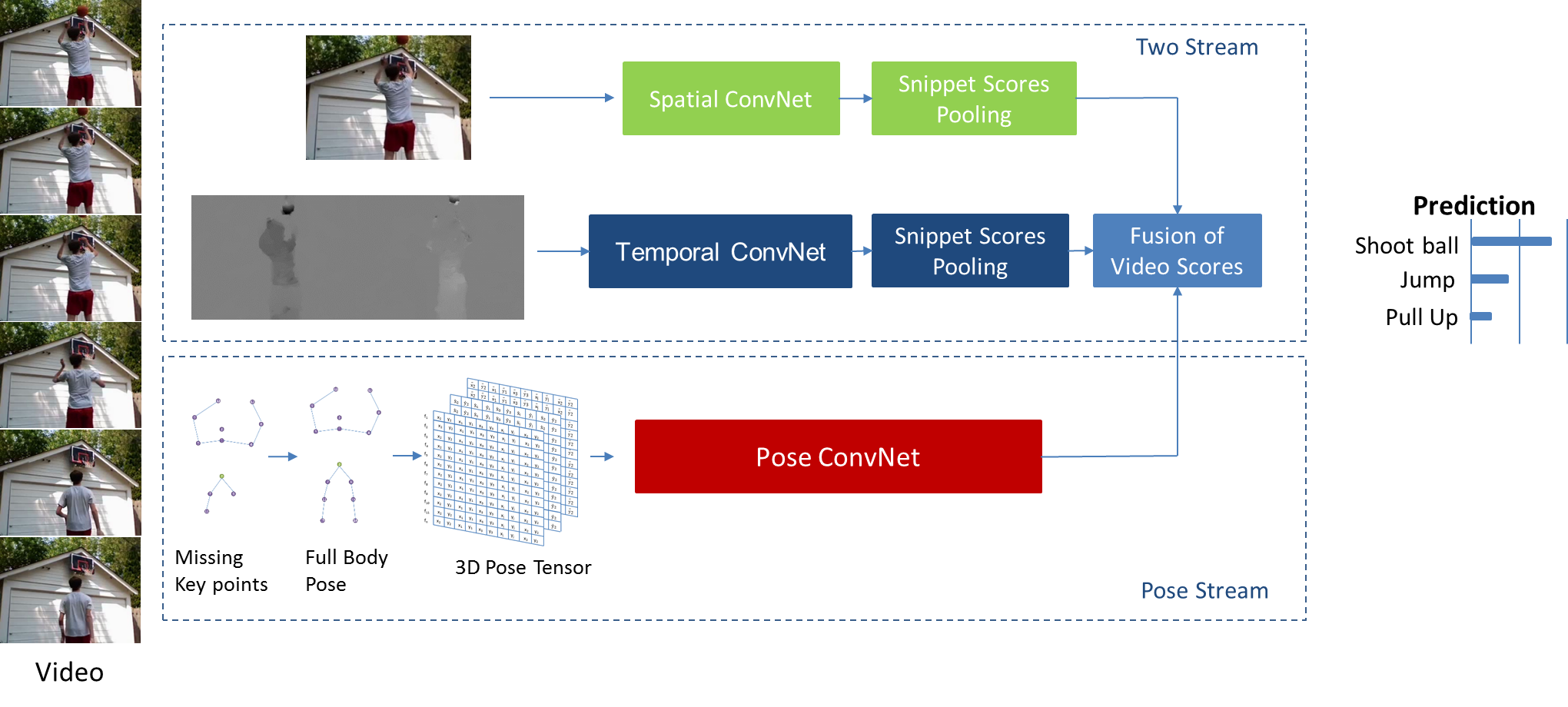}
    \caption{Overview of the proposed three stream network architecture}
    \label{fig_threestream}
  \end{figure*}

\section{Related Work} \label{sec:relatedWork}
Many of the action recognition methods are based on high dimensional features from
videos using hand crafted features. Unsupervised learning approaches like bag-of-words and fisher vectors have been proved to be a very effective way to extract discriminative and compact representation from such high dimensional data \cite{wang2013_ARIT}. Some approaches utilized also deep learning features and combined with hand crafted features as in \cite{wang2015_ARTPDCD}.

To capture temporal structure of actions in video,  \cite{karpathy2014_LSVCCNN} stacked consecutive video
frames and extended the first convolutional layer to learn the spatio-temporal features while exploring
different fusion approaches, including early fusion, slow fusion and late fusion. 
In contrast to previous approaches which can take only fixed number of temporal inputs, \cite{donahue2015_LTRCNVRD} proposed Long Term Recurrent Convolutional Networks
(LRCN) which can work with variable temporal inputs and can also incorporate long term dependencies.
In \cite{wang2016temporal}, a novel sparse spinet concept is proposed to improve the efficiency of temporal sampling by considering the high redundant information between neighboring frames.

Two stream network is proposed in \cite{simonyan2014_TSCNARV} to extract visual and motion features simultaneously, which improved the classification accuracy greatly compared to each feature alone. 
Such an architecture improved many challenging action recognition problems significantly and become more and more popular. In \cite{feichtenhofer2016_CTSNFVAR}, two stream network is exploited with different fusion schemes via 3D convolutional kernels and 3D pooling. 

In addition to the successful two stream networks, human poses are also very popular features utilized to solve human action recognition problems. 
In \cite{wang2013_AAPBAR}, estimated poses are used and coded with bag-of-words approach to classify actions. 
Some approaches like \cite{yao2012_CARPSMV} and \cite{iqbal2017pose} solved pose estimation and action recognition jointly, where \cite{yao2012_CARPSMV} formulated pose estimation as an optimization problem over a set of action specific manifolds and performed two tasks iteratively.
In order to incorporate 3D human poses in CNN, \cite{baradel2017_PCSTAHAR} proposed a novel $3D$ pose-tensor, which preserved the spatial structure of body joints and encoded pose motion in a compact manner. Along with pose CNN, a spatial attention mechanism is used to localize relevant regions for action classification. Inspired by this idea, we propose an extended framework for 2D poses with imperfect joints, so that it can be widely applied in any videos without 3D information available.

  \begin{figure*}[!ht]
    \centering
    \includegraphics[width=6.0in]{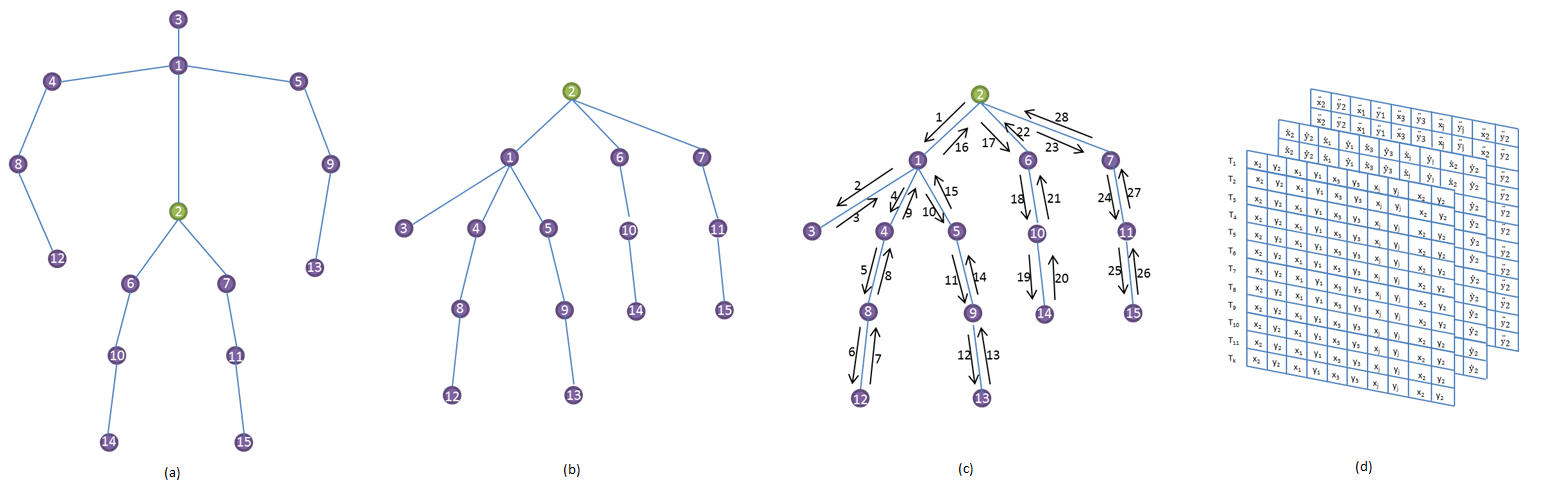}
    \caption{Formation of pose tensor for JHMDB: (a) full body joint position with corresponding labels, (b) tree like structure formed by starting from belly-joint node, (c) traversing of tree like structure by starting from first node and ending at the same node \cite{liu2016spatio} and (d) 3d pose tensor consisting of tree channels, i.e. special ordering of 2d joint positions, the first and second derivative of joint positions}
    \label{fig_pose_tensor}
  \end{figure*}

%%%%%%%%%%%%%%%%%%%%%%%%% Pose Stream %%%%%%%%%%%%%%%%%%%%%%%%%%%%%%%%%
\section{Pose Stream} \label{sec:pose_stream}

In this section, we introduce our technique to use $2D$ pose information of human joints in video for action classification. $2D$ joint positions are arranged in a special formation named pose tensor, that preserves the spatial structure of pose and motion information present in the video. A CNN named Pose ConvNET is trained directly with pose tensor. In contrast to some pose based techniques \cite{baradel2017_PCSTAHAR} that works only with ground truth poses, our technique is robust to work with both ground-truth and detected poses. In our experiments, detected poses are estimated by the 2D CMU pose estimator \cite{cao2017realtime}. However, estimated joint positions are often not completed due to occlusion or other issues in video. We propose two interpolation methods to complete missing joint positions, which improve the training efficiency and performance significantly. Details of the proposed 2D pose tensor and pose ConvNET will be described in the following sections.

  \subsection{Formation of Pose Tensor} \label{sec:poseFormation}
  
    In a video frame, a person can be represented by its $n$ corresponding joints in $2$D image coordinates as shown in Figure \ref{fig_pose_tensor}(a). The joint positions can be either annotated manually, i.e. ground truth \cite{jhuang2013_JHMDB,zhang2013_pennAction} or estimated by a pose estimator \cite{cao2017realtime,wei2016cpm}. Following \cite{liu2016spatio}, a special joint ordering is formulated keeping their neighborhood relationship. Figure \ref{fig_pose_tensor}(b) shows a tree-like structure where each node represents a corresponding joint position. The tree is formed by starting from belly joint and branches are formed with limbs and hand joints as shown in Figure \ref{fig_pose_tensor}(b).
    To form a pose tensor, a path passes from the root node through all subsequent nodes in the pose tree in such a way that all nodes are traversed at least once as shown in Figure \ref{fig_pose_tensor}(c). This traversal keeps the neighborhood relationship among joints preserved in the structure. Based on this path, pose tensor is formed by concatenating all the joint positions $(x,y)$ that occurs in the path traversal in one row of pose tensor as shown in Figure \ref{fig_pose_tensor}(d). Here each row corresponds to the joint positions of one person in one frame. By keeping the same ordering of joints, joint positions in any other frames in a video sequence are stacked row-wisely to form a pose tensor. In this way, a video sequence, corresponding to one action sample, is described by a pose tensor.
    
    More specifically, a video $V$ is divided into $K$ segments $(S_1,S_2,...,S_K)$ of equal length to keep the dimension of pose tensor fixed for all video samples in the dataset,  One snippet $T_k$ is randomly chosen from each segment $S_k$. Then a pose tensor is formed by joint positions of the corresponding person in all snippets $(T_1,T_2,...,T_K)$ as shown in figure \ref{fig_pose_tensor}(d). The second and third channel of pose tensor are the first-order and the second order derivation of joints positions, corresponding to velocity and acceleration of joints in consecutive snippets $(T_{k-1},T_k)$. Thus, $3D$ pose tensor is formulated which not only preserves the spatial structure of human pose but also captures motion information of joints.

%     For forming the pose-tensor, a path connecting the joint-nodes in a certain order using the tree-like structure is formed, see Figure \ref{fig_pose_tensor}(c). The path traverses the tree from the torso-node through each subsequent nodes. It can contain some nodes multiple times assuring that the joint-neighborhood relationship is kept. A row in the pose-tensor, see Figure \ref{fig_pose_tensor}(d), contains a concatenation of the all the joint-nodes pixel coordinates $X$ and $Y$ for $K$ number of frames in a video following the path. This arrangement is similar to \cite{baradel2017_PCSTAHAR}, but using $2$D joint coordinates. The $K$ frames are uniformly taken from the videos independent of their total number of frames. The pose-tensor includes, also, the first $\dot{X}$, $\dot{Y}$ and second moment $\ddot{X}$, and $\ddot{Y}$ of the joint positions. In the end, the pose is represented by a $3$D pose-tensor having the positions, first and second moments concatenated for all the joints following the path for each of the $K$ frames in a video.

  \paragraph{Pose Normalization}\label{sec:poseNorm}
    As $2$D joint positions in image coordinate are sensitive to camera perspectives and image resolution, which are not scale invariant. A normalization is required which keeps all the poses to be of similar size and to be centered in the image. Firstly, joint positions are normalized with respect to torso length which keeps all poses to be of same scale, more specifically as
    \begin{equation}
    {\overline{P_i}}_{(x,y)}=\frac{{P_i}_{(x,y)}}{d} \hspace{2em} \forall i=1,...,n
    \end{equation}
Here, $d$ is the torso length, ${P_i}$ are the raw joint position (x,y), $\overline{P_i}$ are the scaled joint position (x,y) for joint $i$. These scaled joint positions are then shifted by shifting mid-point of torso shifts to origin:
% \begin{equation}
% 	{\overline{P}^{s}_{torso}}_{(x,y)}=\frac{{\overline{P}^{s}_{neck}}_{(x,y)}+{\overline{P}^{s}_{belly}}_{(x,y)}}{2}
% \end{equation}
\begin{equation}
{{PF}_{i}}_{(x,y)}={\overline{P}_{i}}_{(x,y)}-{\overline{P}_{torso}}_{(x,y)} \hspace{1em} \forall i=1,..,n
\end{equation}
Here,${\overline{P}_{torso}}_{(x,y)}$ is the mid-point between neck and belly joint positions, ${{PF}_{i}}_{(x,y)}$ are final normalized joint positions (x,y) for joint $i$ to make 3D pose tensor.%Firstly, joint coordinates are normalized with respect to image coordinate system in the range of $[-1,1]$. Secondly, these are then scaled with respect to torso length. Finally, these normalized joint coordinates are shifted so that the torso joint is located at the origin (0,0). These normalized,  scaled and shifted joint coordinates are used to form a pose tensor.

  \subsection{Handling of Missing Joint Positions} \label{sec:missing}

\begin{figure}[!ht]
      \centering
      \includegraphics[width=3in]{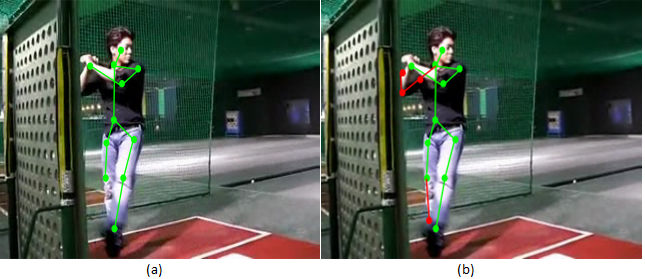}
      \caption{Interpolation of poses (a) original pose detection, (b) completed pose detection after interpolation of joint positions}
      \label{fig_missingPose}
    \end{figure}

      In practice body joints are not always visible in videos, therefore some joint positions cannot be estimated correctly by a pose detector as shown in Figure \ref{fig_missingPose}(a). Handling of such missing joint positions is a critical part by formulating pose tensor. Simply marking these points as invalid or assigning a specific value, would corrupt the input and cause some unexpected issues by training a CNN on the pose tensor. Therefore, two interpolation techniques are implemented to estimate missing joint positions: temporal interpolation using the joint positions available in other frames, and spatial interpolation exploiting spatially neighbored existing joint in the same frame.

%     \begin{figure}[!ht]
%        \centering
%        \includegraphics[width=3in]{complete_pose_spatial_temporal.png}
%        \caption{Interpolation of Missing Key points}
%        \label{fig_interpolatedPose}
%      \end{figure}

	\subsubsection{Temporal Interpolation of Pose}

	Videos contain rich motion information covering the smooth movement of human joints. One consequence is that some invisible joints become visible in the continuous frames or inversely. By making using of the continuity of the joint movement, the position of invisible joints can be interpolated from temporally neighbored visible joints if any. We used a simple linear interpolation to estimate location of missing joint positions which achieves promising estimation for short temporal range. 
	Temporal interpolation is especially useful by estimating missing joint positions with short-termly changing visibility.

  \subsubsection{Spatial Interpolation of Pose}
  For long-term occluded human joints, temporal interpolation has its limitation. If some joints are not detected for a long temporal range, the linear motion assumption made by temporal interpolation is not valid any more. Therefore, we exploit spatial context information of neighbored joints to estimate the missing joint. This idea is based on the fact that locations of joints of each pose are strongly statistically correlated, especially among neighbored joints, e.g. head and shoulder. Similar as \cite{muller2010human}, neighborhood relationships between joints are utilized to vote possible location of missing joints. A polynomial function is used to model the spatial relationship of neighbored joints. This model is learned from varied video datasets with ground-truth poses.
  
\begin{figure}[ht]
        \centering
        \includegraphics[width=1.2in]{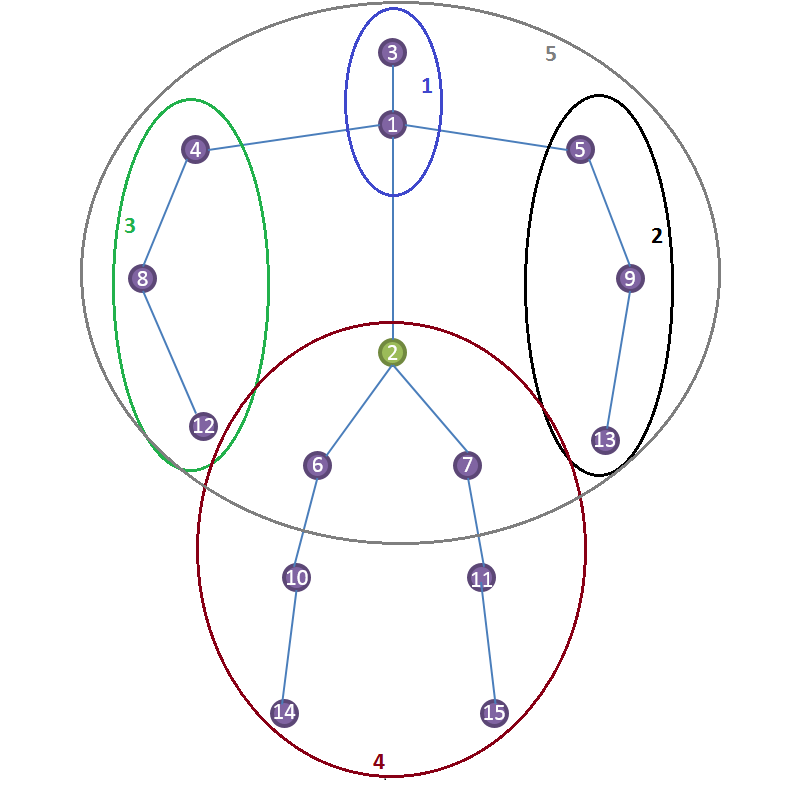}
        \caption{Configuration of the $5$ body parts for spatial interpolation of missing joint positions}
        \label{fig_spatialInterpolation}
      \end{figure}

  As directly neighbored joints provide more accurate estimation, the whole body is divided into 5 body parts keeping their tight neighborhood relations of joints as shown in Figure \ref{fig_spatialInterpolation}, where part 1 to part 4 have tight spatial relationship and part 5 has only a loose spatial relationship.
  For a missing joint position within frame, first all available joint positions of the corresponding body part with tight spatial relationship, i.e. part 1 to part 4, are selected to estimate the missing joint position. If no joint position from the corresponding body part is available, then joints from body part 5 are selected for missing upper body joints. For other cases, all the available joint positions of this pose in that frame are selected. Each selected joint position votes for the position of the missing joint and the average vote of all selected joints is considered as the final estimation of the missing joint.

  \subsection{Pose ConvNET} \label{sec:architecture}
  A CNN (Pose ConvNET) is trained in an end-to-end fashion with Pose Tensor. This ConvNET have two convolutional layers with a RELU function along with Max Pooling Layer. Final features are extracted via a fully connected layer and a fully connected softmax layer is used for classification. A relative shallow network is used with small filter size $3\times 2$, as pose tensor is of highly compact data and consists of only high level features. An important advantage is that neither large amount of training data nor properly pre-training are needed, therefore a flexible training for varied real-world applications is possible. The network is trained with Xavier initialization and a standard categorical cross-entropy loss function. A Pose ConvNET trained for a video $V$ with $K$ segments can be mathematically defined as:
  
  \begin{equation}
  PCN(V)=F((T_1,T_2,...,T_K);W)
  \label{eq: PCN}
  \end{equation}
  Here, $F$ is the function representing Pose ConvNET with parameters $W$ which operates on pose tensor $(T_1,T_2,...,T_K)$ formed by $K$ snippets and produces class scores for video $V$.

%       We introduce the CNN architecture that is trained using pose-tensors. This pose CNN contains two blocks, where each of them have a convolutional layer with ReLU activation function followed by max-pooling. Besides, it has a fully-connected layer, and a softmax layer as a classifier. The network predicts action classes for videos using pose-tensors as data sources. The networks is shallow, as the pose-tensors include high-level pose features. The convolutional layer contain $3\times 2$ filters, which extract features from corresponding neighboring joints.

%%%%%%%%%%%%%%%%% Three-Stream Convolutional Neural Network %%%%%%%%%%%%%%%%%%%%%%%

\section{Three-Stream Convolutional Neural Network} \label{sec:threeNet}

Fusion of two stream CNN \cite{wang2016temporal, simonyan2014_TSCNARV,feichtenhofer2016_CTSNFVAR} based on RGB and optical flows has given promising results in the domain of human action recognition. We extend this framework by fusing an additional stream of pose tensor with the conventional two stream CNN. A three stream network is designed to capture context information from a spatial channel, motion information from a flow channel and semantic posture information using the proposed pose ConvNET.

We use the TSN framework proposed in \cite{wang2016temporal} for training of RGB and optical flow streams, where warped flow fields \cite{wang2013_ARIT} are calculated to compensate camera motion, to suppresses background motion and to make motion concentrated on actors, similar as a visual saliency map.
Pre-trained spatial and temporal models on videos of UCF101 \cite{soomro2012ucf101} are used and fine-tuned on the new datasets.

Following the sampling concept proposed in TSN framework, each video $V$ is divided into $P$ equal segments $(S_1,S_2,..,S_P)$. For each segment, one snippet $T_p$ for spatial stream and stack of consecutive snippets within segment $S_p$ for temporal stream are randomly sampled. Each CNN model is trained separately with these sampled snippets from video. The temporal segment network is defined mathematically as
\begin{equation} \label{eq. TSN}
TSN(V)=G(F(T_1;W),F(T_2;W),....F(T_P;W))
\end{equation}
Here, $F(T_p;W)$ is the function representing spatial and temporal CNN models with parameters $W$ which operates on the short snippet $T_p$ and produces class scores for that snippet. $G$ represents the segmental consensus function which aggregates the scores from all the snippets within one video and gives video based score. We used average pooling of scores as consensus function $G$. During training of each of TSN streams, this aggregated video level prediction is used to minimize the loss function and errors are propagated through back propagation algorithm. 

Three stream convolutional neural network (TSCNN) is formulated by fusing scores from Pose ConvNET with video based scores from spatial and temporal streams of TSN. The final score will be weighted sum of scores as given below:
% \begin{equation} \label{eq. TSCNN}
% TS_{CNN}(V)= \\
% argmax(w_p*PCN(V)+w_s*TSN_s(V)+w_t*TSN_t(V))
% \end{equation}
\begin{equation}
%\resizebox{.45 \textwidth}{!} 
\begin{split}
& TSCNN(V) = \\
& w_p*PCN(V)+w_s*{TSN}_s(V)+w_t*{TSN}_t(V),
\end{split}
\end{equation}
where $PCN$, $TSN_s$ and $TSN_t$ are video based scores for pose ConvNET, spatial  and temporal streams as shown in Figure \ref{fig_threestream}. $w_p$, $w_s$ and $w_t$ are the weights accordingly, which are estimated empirically.

    \begin{table}[!ht]
      \caption{Classification accuracy for each of models trained on RGB, optical flow, poses (ground truth, estimated) of the JHMDB, subJHMDB and Penn Action datasets}
      \label{table:JHMDB}
      \begin{center}
      \begin{tabular}{|c | c | c | c |}
        \hline
        \multirow{2}{*}{\textbf{Network}}	& \multicolumn{3}{c|}{\textbf{Accuracy}}	\\
        \cline{2-4}
        & JHMDB & subJHMDB  & Penn Action \\
        \hline
        Spatial (RGB frame) 				& 	57.90\%		&	58.76\%  & 86.42\%	\\ 
        Temporal (Optical flow)			&	73.33\%		&	81.14\%	 & 96.72\%	 \\ 
        Pose (GT pose)				& 	70.84\%		&	75.44\%	 &  96.25\% \\  
        Pose (Est. pose)		&	54.90\%		&	63.60\%	 &	89.32\%\\ 
        \hline

      \end{tabular}
      \end{center}
    \end{table}

%%%%%%%%%%%%%%%%%%%%%%%%%%%% Experiments %%%%%%%%%%%%%%%%%%%%%%%%%%%%%%%%%%%%
\section{Experiments} \label{sec:experiments}
In this section, experiments on datasets JHMDB, sub-JHMDB \cite{jhuang2013_JHMDB} and Penn Action \cite{zhang2013_pennAction} are presented along with some specific
implementation details. All
datasets contain varied action videos with action labels and 2D human pose annotations,
which are required by the Pose ConvNET stream. We explore the
performance of each stream and their fusion in terms of accuracy of action
classification. Finally we
compare the performance of our approach to some state-of-the-art approaches.
% 	We have evaluated the aforementioned pose CNN and the three-stream framework on three datasets: JHMDB, sub-JHMDB \cite{jhuang2013_JHMDB} and Penn Action \cite{zhang2013_pennAction}. Datasets contain videos for HAR, including, as well, $2$D annotations of human joint pixel coordinates. We explored the performance of each data source, and their fusion in terms of accuracy of action classification.

  \subsection{JHMDB} \label{sec:JHMDB}
    JHMDB dataset \cite{jhuang2013_JHMDB} contains $21$ action classes with total of
$928$ videos and $33183$ frames. A subset of the JHMDB named sub-JHMDB is also provided with $316$ videos and $12$ action classes. Different environments, changing camera view points and high intra-variations of actions are covered in both datasets. All joints are annotated manually even under occlusion. We conducted two experiments: one is based on $2D$ annotated joint positions (GT Pose) with $n=15$, another one is based on estimated $2D$ joint positions (Est. Pose) by \cite{cao2017realtime} with $n=14$, where 4 face key points are discarded. All joint positions are normalized and missing joint positions in case of estimated poses are estimated by applying first the temporal interpolation and then the spatial interpolation (see Section \ref{sec:pose_stream}). From them the final pose tensor is formed with $K=15$ and of size ($15\times 58 \times 3$) for GT Pose and ($15\times 54 \times 3$) for Est. Pose. 
% We constructed two pose-tensors with $n=15$ joints, the neck joint as root node, $K=15$ and $F=5$ per video, having a final size of ($15\times 58 \times 3$), see Section \ref{sec:poseFormation}. One pose-tensor is built utilizing the annotated joint positions (GT pose) and the other with estimated joint positions (Est pose), see Section \ref{sec:missing}. A $3$-fold cross-validation method is carried out; that is, the dataset is split in $3$ parts, two used for training and the remaining one for testing.

For comparison, four CNN models are trained separately on each cue, i.e. RGB, optical flow and poses (GT and Est). Pre-trained spatial and temporal models on UCF101 \cite{soomro2012ucf101} are used with $P=15$ snippets and their fully connected layers are fine tuned with JHMDB and sub-JHMDB datasets. According to the standard protocol, three splits are provided. Experiments are performed on all three splits and averaged results are reported in Table \ref{table:JHMDB}. The temporal model performs best on JHMDB and sub-JHMDB. In contrast, the spatial model is much less performing. It shows that motion is much more important feature than image context on both datasets, that matches our observations as well. The model trained with GT Pose shows close performance to the temporal model. However, the model trained with Est. Pose has a significant accuracy drop, especially on JHMDB, where full bodies are often not visible, which decreases the performance. It shows that our interpolation methods have some limitations by facing lots missing joint positions in video.
    
%     First, we have trained four CNNs for each of the clues: RGB, optical flows, and the two pose-tensor (GT and Est). The spatial and temporal networks are pre-trained with the large video dataset UCF101 \cite{soomro2012ucf101}, and then fine-tuned their fully-connected layers using the three training splits. Evaluations are carried out with the three testing splits and their performances are averaged. Table \ref{table:JHMDB} shows the averaged accuracy on the four CNNs on the three testing splits. The temporal CNN trained with Optical flows shows the best results, due to the more quantity of motion related videos in the dataset. Moreover, the pose CNN on GT pose-tensors presents higher accuracy in comparison with the spatial network on RGB frames. The pose CNN on estimated pose-tensor shows the worst accuracy because for most of the videos, full person was not available and interpolation of missing joint positions doesn't work well in this situation. However, its performance is not far from the spatial network. Similarly, the experiments are performed on the three splits of sub-JHMDB. Table \ref{table:JHMDB} shows the average classification accuracy of the three splits. These results follow the same behavior of the biggest dataset.

  \subsection{Penn Action Dataset}
  
    The Penn Action dataset contains $15$ action categories. The dataset provides both action labels and positions of $n=13$ human joints even under occlusion. Following the setting in \cite{zhang2013_pennAction}, data are divided into 50/50 for training and testing.
    
The spatial, temporal and pose models are trained similarly as in Section \ref{sec:JHMDB}. Results of each individual stream: RGB, optical flows and pose tensor (GT and Est.) are reported in Table \ref{table:JHMDB}. Pose tensor based on GT pose was built with $n=13$ joints, head joint as root node and $K=15$ snippets. Thus, the size of pose tensor with GT Pose was $(15\times 50\times 3)$. Similar trends can be observed as that on JHMDB, where the temporal model performs best. However, the results of the model trained on GT Pose are very close to that of temporal model. The model trained on estimated poses performs better than spatial model, despite the fact that pose tensor has more compact input. It shows the power of semantic features by learning. 
    
%     For training and testing the spatial- and temporal- CNNs, we used similar settings as described in Section \ref{sec:JHMDB}. Table \ref{table:JHMDB} also shows the testing accuracy for classifying videos into action classes for each individual stream: RGB, optical flows and pose-tensors with GT and estimated human joint positions. For training and testing of the pose CNN, pose-tensors per video are built with $n=13$ joints, the head joint as the root node, $K=15$ frames, $F=5$, see Section \ref{sec:pose_stream}. Pose tensors are of size $(15\times 50\times 3)$. The temporal CNN presents the best results, as with the JHMDB. However, the pose CNN on GT pose-tensors is only $0.47\%$ lesser than the temporal CNN, and the pose CNN on Est pose-tensors performs better than the spatial CNN. These comparisons demonstrate that the usage of human body information is a suitable data source for training deep architectures for HAR.

      \begin{table}[!ht]
      	\scriptsize
        \caption{Classification Accuracy for each of the spatial, temporal and pose CNNs on the JHMDB and sub-JHMDB datasets}
        \label{table:fusion_JHMDB}
        \begin{center}
        \begin{tabular}{|c | c | c | c | c |} \hline
            \multirow{3}{*}{\textbf{Fusion Modularity}}	&	\multicolumn{4}{c|}{\textbf{Accuracy}}\\
            \cline{2-5}
            &  \multicolumn{2}{c|}{\textbf{JHMDB}} & \multicolumn{2}{c|}{\textbf{sub-JHMDB}}	\\
            & 	Pose (GT) 	&	Pose (est.) & 	Pose (GT) 	&	Pose (est.)	\\
            \hline
            \ RGB + flow 		& 	\multicolumn{2}{c|}{75.83\%} & \multicolumn{2}{c|}{78.09\%} \\
            % \ RGB + flow 		& 	75.83\%  & 75.83\%  &	78.09\% & 78.09\% \\
            \ RGB + pose		&	73.45\%	 &	62.86\%	 &  69.02\% & 66.10\%	\\
            \ flow + pose		& 	79.32\%	 &	71.69\%	 &  83.20\% & 81.30\%	\\
            \ RGB + flow + pose	&	83.05\%	 & 	78.81\%	 & 87.29\%  & 85.12\%	\\
            \hline
        \end{tabular}
        \end{center}
      \end{table}
      
      \begin{table}[!ht]
        \caption{Classification Accuracy for each of the spatial, temporal and pose ConvNETs on Penn Action dataset}
        \label{table:fusion_Penn}
        \centering
        \begin{tabular}{|c | c | c|} \hline
            \multirow{2}{*}{\textbf{Fusion Modularity}}	&	\multicolumn{2}{c|}{\textbf{Accuracy}}\\
            \cline{2-3}
            & 	pose (GT) 	&	pose (Estimated)	\\
            \hline
            RGB + flow 			& 	\multicolumn{2}{c|}{95.04\%} \\
            RGB + pose			&	93.72\%		&	91.67\%			\\ 
            flow + pose			& 	97.85\%		&	97.10\%			\\ 
            RGB + flow + pose		&	98.50\%		& 	98.41\%			\\
            \hline
        \end{tabular}
      \end{table}

  \subsection{Fusion of Multiple Cues}
  In this section varied fusion schemes are evaluated on the JHMDB, sub-JHMDB and Penn Action Datasets. Table \ref{table:fusion_JHMDB} shows the performance on the JHMDB, sub-JHMDB of four different combination of three cues, RGB + optical flow (conventional two stream network), RGB + pose, flow + pose, and RGB + flow +pose, with two pose variants, GT and estimated pose. For all the experiments, we used $(w_p,w_s,w_t)=(1,1,1)$ as the weights for fusion of three streams. Comparing to conventional two stream fusion configurations proposed in \cite{wang2016temporal}, improvements of $7.2\%$ and $9.2\%$ respectively are achieved on the JHMDB and sub-JHMDB by using GT pose, while $2.98\%$ and $7.03\%$ by using estimated poses. Even the fusion of the temporal and pose models outperforms the conventional two stream. A clear benefit by fusing additional pose feature can be observed.
  
	Similar results on the Penn action dataset are shown in Table \ref{table:fusion_Penn}: the performance of the three stream network using the GT human pose is $3.46\%$ better than the RGB and optical flow fusion, proposed in \cite{wang2016temporal}. Even the fusion using estimated poses is very close to three stream with GT pose. It shows that recent pose estimators are already very stable on some real world data.

      \begin{table}[!ht]
      	\scriptsize
		\caption{Comparison of the three stream CNN (TSCNN) with estimated (Est) and Ground-truth (GT) human joint positions with the State-of-the-art on JHMDB, sub-JHMDB and Penn Action Datasets.}
        \label{Table : Comparison with state-of-the-art}
        \centering
        \begin{tabular}{|c | c  c  c |}
        	\hline
            State-of-the-art						&	JHMDB	& 	Sub-JHMDB 	&	PennAction 		\\
            \hline
            Pose\cite{jhuang2013_JHMDB}				& 	69	&	52.9 &	-  \\ 
            STIP\cite{zhang2013_pennAction} 		& 	-	&	- &	82.9 \\ 
            Action Bank\cite{zhang2013_pennAction}	&	-	&	- &	83.9 \\
            MST\cite{wang2014cross} 				& 	- 	& 45.3 & 74.0	 \\ 
            AOG\cite{xiaohan2015joint}				&	-	&	61.2 & 85.5	\\ 
            P-CNN\cite{cheron2015pcnn}				&	79.5 &	72.5 &	-	\\
            Hierarchical\cite{lillo2016hierarchical}&	-	&	77.5 &	-	\\
            IHLPF \cite{fan2016NHLP}				&	80.4 &	- &	- \\ 
            JDD\cite{cao2016action}					&	-	&	83.3 &	95.7 \\
            Pose+idt-fv\cite{iqbal2017pose}			&	-	&	74.6 &	92.9 \\
            RPAN-(S+T)\cite{du2017rpan}				&	-	&	81.1 &	97.4 \\
            \hline
            TSCNN Est pose (Our)	&	78.8 &	85.1 &	98.4	\\
            TSCNN GT pose (Our)       &	\textbf{83.1} &	\textbf{87.3} &	\textbf{98.5}	\\
            \hline
        \end{tabular}
      \end{table}
\subsection{Comparison to State-of-the-art Approaches}
	A comparison of our three stream network using GT and estimated joint positions with recent state-of-the-art deep learning and conventional hand crafted approaches for JHMDB, sub-JHMDB and Penn Action datasets are reported in Table \ref{Table : Comparison with state-of-the-art}. Clearly, apart from JHMDB with estimated Poses, our proposed three stream network outperforms the recent state-of-the-art approaches with significant difference for all three datasets, with GT and Estimated Poses. On JHMDB the three stream network with estimated poses has a lower performance due to frequently invisible body parts as mentioned in the previous section. These results explain that our proposed fusion scheme of three cues shows a complementary behavior.
% 	Table \ref{Table : Comparison with state-of-the-art} shows a comparison among our three stream network with the GT and estimated human joint positions, and other recent deep learning and hand-crafted features based approaches on JHMDB, Sub-JHMDB and Penn Action datasets. In general, the performance of the proposed three stream network is superior with respect to the other approaches. On JHMDB the three stream network with estimated poses has a lower performance due to frequently invisible body parts as mentioned in the previous section. These results demonstrate that our proposed fusion scheme of three cues exhibits a complementary behavior.
    
%       \begin{table}[!ht]
%       	\tiny
%         \caption{Classification Accuracy for each class of subJHMDB split1 for each cue and their fusion}
%         \label{table:analysis}
%         \centering
%         \begin{tabular}{|c | c | c| c | c | c | c| c | c | c |} \hline
%             \multirow{2}{*}{\textbf{Networks}}	&	\multicolumn{9}{c|}{\textbf{Actions}}\\
%             \cline{2-10}
%             & 	Catch  & Golf & Jump & Kick ball & Pick & Push & Run & Shoot ball & Walk \\
%             \hline
%             Spatial	& 0.4 & 1.0 & 0.25 & 0.63 & 0.63 & 1.0 & 0.43 & 0.17 & 0.5 \\
%             Temporal& 0.6 & 1.0 & 0.75 & 0.88 & 1.0 & 1.0 & 0.71 & 0.33 & 0.5 \\ 
%             Pose	& 0.4 & 1.0 & 0.63 & 0.38 & 1.0 & 0.9 & 0.43 & 1.0 & 1.0 \\ 
%             Fusion	& 0.8 & 1.0 & 0.88 & 0.75 & 1.0 & 1.0 & 0.71 & 0.83 & 1.0 \\
%             \hline
%         \end{tabular}
%       \end{table}
\begin{figure}[!ht]
      \centering
      \includegraphics[width=2.5in]{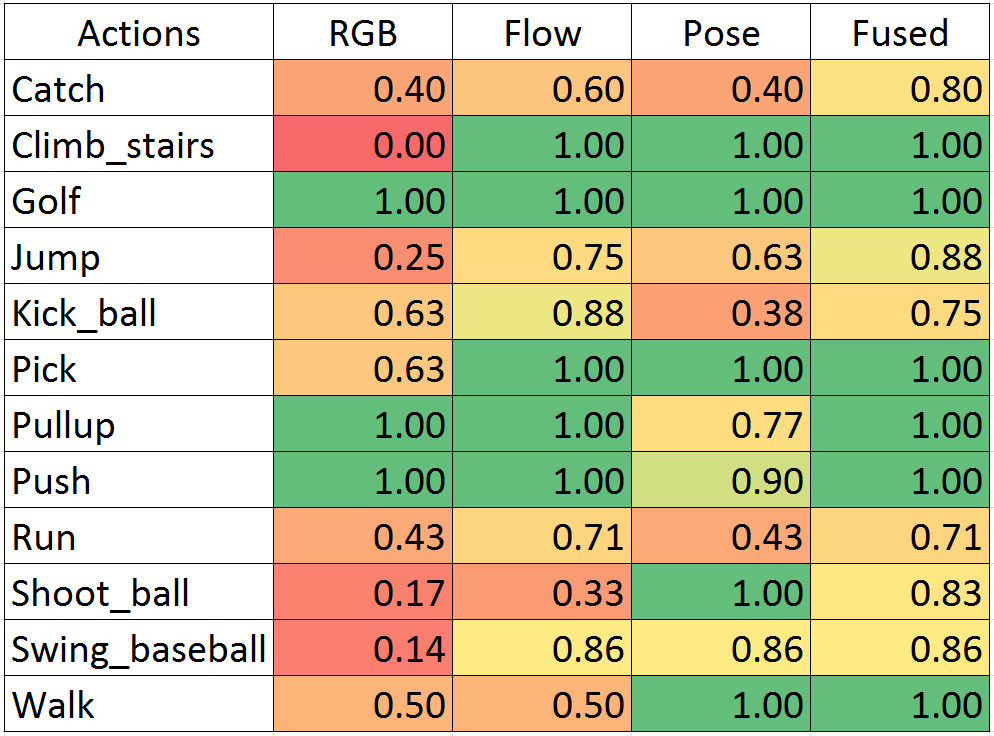}
      \caption{Classification Accuracy of each classs on subJHMDB split 1 for each cue and fusion}
      \label{fig_qualititative analysis}
    \end{figure}
    
\subsection{Qualitative Analysis of Cues}
In order to get more insights of the complementary behavior of different cues, some examples are qualitatively examined and summarized in Figure \ref{fig_qualititative analysis}. It is clear that no single cue alone gets an overall good performance on varied action classes, as the fused cues do. It is observed that the flow cue works especially good on actions with fast motions, e.g. run and swing baseball, while the pose cue contributes much to actions with unique posture or significant body motion, e.g. climb stair, pick and shoot ball. The RGB cue performance worse than other two cues, however it is still very important by understanding the context information, as meadow for action "golf". It is confirmed that almost all actions are improved by fusing all cue together. However, it is not a trivial task to identify contribution of each cue on different actions empirically. How to learn the fusion scheme dynamically, is an important research topic for the future.
      
  %\subsection{Qualitative Analysis of Different Cues} 
      %Table \ref{fig_qualitative_analysis} shows the classification accuracies of some classes of the .. dataset for the three CNNs and their fusion. For some action classes like "Walk" and "Shoot ball", pose is the vital cue which contributed the most in correct classification. For action classes such as "Run" and "Kick ball", flow is the most efficient cue. By using our Fusion method, clearly merits of each individual cue is utilized and classification accuracy of each class gets improved.

%       \begin{table}[!ht]
%       	\footnotesize
% 		\caption{Classification Accuracy for each class of subJHMDB split1 for each cue and their fusion}
%         \label{Table : Comparison with state-of-the-art}
%         \centering
%         \begin{tabular}{|c  c  c  c c  c  c  c c  c  c  c c|}
%         	\hline
%             Actions & Catch & Climb Stairs & Golf & Jump & Kick ball & Pick & Pullup & Push & Run & Shoot ball & Swing baseball & Walk \\
%             \hline
%         \end{tabular}
%       \end{table}

%    \begin{figure}[!ht]
%      \centering
%      \includegraphics[width=2.5in]{CueAnalysis-subJHMDB1.PNG}
%      \caption{Classification Accuracy for each class of subJHMDB split1 for each cue and their fusion}
%      \label{fig_qualitative_analysis}
%    \end{figure} 

%%%%%%%%%%%%%%%%%%%%%%%%%%%% Conclusion %%%%%%%%%%%%%%%%%%%%%%%%%%%%%%%%%%%%
\section{Conclusion} \label{sec:conclusion}
	This paper has presented a novel framework to utilize human body poses along with RGB frames and optical flows for action recognition. Both GT and estimated poses are supported, that enables a wide range of applications in real world. In experiments, very promising results are shown in the benchmarking datasets and outperform recent state-of-the-art approaches. The complementary behavior of RGB, optical flow and pose is observed and analysed in our experiments. Dynamic adaptation of fusion scheme for different actions will be investigated in the future.
    
    %Further improvement in relation with the estimation and interpolation of human joints, and more elaborated fusion schemes can be carried out. 

\section*{Acknowledgment}
This work was supported by the Computer Vision Research Lab of Robert Bosch GmbH and Fraunhofer IPA.



%\bibliographystyle{IEEEtran}

%%%%%%%%%%%%% BIBLIOGRAPHY %%%%%%%%%%%%%%%%%%%%%%%%%%%%
% Add the bibliography entries in literatur.bib

\bibliographystyle{IEEEtran}
\bibliography{bibliography}

\end{document}